\documentclass{article}
\usepackage{modifiedNIPS15,times}
\usepackage{hyperref}
\usepackage{url}
\usepackage{amsfonts}
\usepackage{amsmath}
\usepackage{varioref}
\usepackage{amssymb}
\usepackage{bm}
\usepackage{color}
\usepackage{graphicx}
\usepackage{booktabs}

\DeclareMathOperator*{\argmin}{arg\,min}

\DeclareMathOperator{\st}{subject\:to}

\DeclareMathOperator{\Trc}{Tr}

\def\Xiz{{\boldsymbol{\Xi}}}

\def\Omegaz{{\boldsymbol{\Omega}}}
\def\Lambdaz{{\boldsymbol{\Lambda}}}

\def\Sigmaz{\boldsymbol{\Sigma}}

\newcommand{\UU}{{\mathbf U}}
\newcommand{\VV}{{\mathbf V}}
\newcommand{\WW}{{\mathbf W}}
\newcommand{\CC}{{\mathbf C}}
\newcommand{\XX}{{\mathbf X}}
\newcommand{\YY}{{\mathbf Y}}
\newcommand{\II}{{\mathbf I}}
\newcommand{\MM}{{\mathbf M}}
\newcommand{\QQ}{{\mathbf Q}}

\newcommand{\PP}{{\mathbf P}}

\newcommand{\RR}{{\mathbf R}}
\def\uz{{\boldsymbol{u}}}

\def\xz{{\boldsymbol{x}}}
\def\yz{{\boldsymbol{y}}}

\title{Why (and How) Avoid Orthogonal Procrustes in Regularized Multivariate Analysis}

\author{
Sergio~Mu\~noz-Romero, Vanessa~G\'omez-Verdejo,  Jer\'onimo~Arenas-Garc\'ia \\
Signal Processing and Communications\\
Universidad Carlos III de Madrid\\
28911 Legan\'es (Madrid), Spain \\
\texttt{smunoz,vanessa,jarenas@tsc.uc3m.es}
}

\nipsfinalcopy 

\begin{document}

\maketitle

\begin{abstract}
Multivariate Analysis (MVA) comprises a family of well-known methods for feature extraction that exploit correlations among input variables of the data representation. One important property that is enjoyed by most such methods is uncorrelation among the extracted features. Recently, regularized versions of MVA methods have appeared in the literature, mainly with the goal to gain interpretability of the solution. In these cases, the solutions can no longer be obtained in a closed manner, and it is frequent to recur to the iteration of two steps, one of them being an orthogonal Procrustes problem. This letter shows that the Procrustes solution is not optimal from the perspective of the overall MVA method, and proposes an alternative approach based on the solution of an eigenvalue problem. Our method ensures the preservation of several properties of the original methods, most notably the uncorrelation of the extracted features, as demonstrated theoretically and through a collection of selected experiments.

\end{abstract}

\section{Introduction}
MultiVariate Analysis (MVA) techniques have been widely used during the last century since Principal Component Analysis (PCA) \cite{pearson1901pca} was proposed as a simple and efficient way to reduce data dimension by projecting the data over the maximum variance directions. Since then, many variants have emerged trying to include supervised information to this data dimension reduction process; this is the case of algorithms such as Canonical Correlation Analysis (CCA) \cite{hotelling1936cca}, Partial Least Squares (PLS) approaches \cite{wold1966nipals2,wold1966nipals1}, or Orthonormalized PLS (OPLS) \cite{worsley1998mvlm}. In fact, we can find many real applications where these methods have been successfully applied: in biomedical engineering \cite{Gerven12,Hansen07}, remote sensing \cite{Arenas08,Arenasbook}, or chemometrics \cite{Barker03}, among many others.

Some recent significant contributions in the field have focused on trying to gain interpretability by means of including $\ell_1$ and $\ell_{2,1}$ norms, or even group lasso penalties, in the MVA formulations. This is the case of extensions such as sparse PCA \cite{Zou06}, sparse OPLS \cite{Gerven12}, group-lasso penalized OPLS (or SRRR) \cite{Chen12}, and $\ell_{2,1}$-regularized CCA (or L21SDA) \cite{Shi14}. All these approaches are based on an iterative process which combines two optimization problems. The first step consists of a regularized least-squares problem to obtain the vectors for the extraction of input features; the second step involves a minimization problem which is typically solved as an orthogonal Procrustes problem.

Despite these regularized approaches have been recurrently applied in feature extraction and dimensionality reduction scenarios \cite{Gerven12}, \cite{Shi14}, all of them ignore one intrinsic and important property of most MVA approaches: \textit{uncorrelation of the extracted features in the new subspace}. When this property holds, the feature extraction process provides additional advantages: (1) The subsequent learning tasks (working over this new space) are easen, for instance, least-square problems (Ridge Regression, LASSO,...) can work independently over each dimension, and the effects of variations of the input data are isolated in the different directions. (2)  The selection of optimal feature subsets becomes straightforward; since once a set of features is computed, obtaining an optimum reduced subset consists of selecting those features with highest associated eigenvalue. Consequently, the adjustment of the optimum number of extracted features is simplified.


In this paper, we analyze in detail the above mentioned MVA formulations showing, from a theoretical and experimental point of view, some drawbacks overlooked  until now in the literature. Concretely, we will demonstrate that these MVA approaches (1) do not obtain uncorrelated features in general; (2) do not converge to their associated non-regularized MVA solutions; (3) suffer some issues that depend on the algorithm initialization, e.g., depending on algorithm initialization the methods can fail to progress at all.

As solution to these problems, this paper proposes an alternative to orthogonal Procrustes. In order to do so, we rely on a common framework that allows us to deal simultaneously with the most common MVA methods (PCA, CCA and OPLS), and extend it to favor interpretable solutions by including a regularization term. Similarly to existing methods, we propose a solution to this generalized formulation which is based on an iterative process but does not suffer from the above problems.

The paper is organized as follows: Firstly, Section 2 introduces this generalized MVA framework. Then, Section 3 presents the iterative process required to solve its regularized extension and  describes both Procrustes solution, as well our proposal based on a standard eigenvalue problem. Section 4 explains the limitations of the Procrustes solution in greater detail, and provide theoretical proof of the most important problems of this approach. Section 5 illustrates and compare the suitability of the new proposed solution with that of Procrustes using some real problems that support well the theoretical findings. Finally, Section 6 concludes the paper.

\section{Framework for MVA with uncorrelated features}

This section reviews some well-known MVA methods under a unifying framework, so that subsequent sections can deal with these methods in a unified manner. Before that, notation used throughout the paper is presented. 

Let us assume a supervised learning scenario, where the goal is to learn relevant features from an input data set of $N$ training data $\{\xz_i,\yz_i\}_{i = 1}^N$, where $\xz_i \in \Re^n$ and $\yz_i \in \Re^m$ are considered as the input and output vectors, respectively. Therefore, $n$ and $m$ denote the dimensions of the input and output spaces.  For notational convenience, we define the input and output data matrices: $\XX = \left[\xz_1,\dots,\xz_N \right]$ and $\YY = \left[\yz_1,\dots,\yz_N \right]$, so sample estimations of the input and output data covariance matrices, as well as of their cross-covariance matrix, can be calculated as $\CC_{\XX\XX} = \XX\XX^\top$, $\CC_{\YY\YY} = \YY\YY^\top$ and $\CC_{\XX\YY} = \XX\YY^\top$, where we have neglected the scaling factor $\frac{1}{N}$, and superscript $^\top$ denotes vector or matrix transposition. The goal of linear MVA methods is to find relevant features by combining the original variables, i.e., $\XX' = \UU^\top \XX$, where the $k$th column of $\UU = [\uz_1,\dots,\uz_{n_f}]$ is a vector containing the coefficients associated to the $k$th extracted feature.

The results in this paper apply, at least, to PCA, CCA, and OPLS, all these methods having in common that the extracted features are uncorrelated, i.e., $\UU^\top \CC_{\XX\XX} \UU = \Lambdaz$, with $\Lambdaz$ a diagonal matrix. MVA methods that do not enforce this uncorrelation, more notably PLS, are therefore left outside the scope of this paper.

A common framework for many regularized MVA methods can be found in \cite{reinsel98}. According to it, these methods pursue the minimization of the following objective function:
\begin{align}
\label{GOPLS_cost}
{\cal L}(\WW,\UU) &= \|\Omegaz ^{\frac{1}{2}} \left(\YY  - \WW \UU^\top \XX\right) \|_F^2 + \gamma R\left(\UU\right) \nonumber\\
 &= \Trc\{\YY^\top\Omegaz \YY\} - 2 \Trc\{\UU^\top \CC_{\XX\YY}\Omegaz \WW\} + \Trc\{\UU^\top \CC_{\XX\XX} \UU \WW^\top \Omegaz  \WW\} + \gamma R\left(\UU\right),
\end{align}
where $R\left(\UU\right)$ is a regularization term, such as the ridge regularization ($||\UU||^2$), the $\ell_1$ norm ($|\UU|_1$), or the $\ell_{2,1}$ penalty for variable selection ($||\UU||_{2,1}$). Parameter $\gamma$ trades off the importance of the regularization term in \eqref{GOPLS_cost}, $\WW$ can be considered a matrix for the extraction of output features, and different particularizations of matrix $\Omegaz$ give rise to the considered MVA methods, in particular $\Omegaz=\CC_{\YY\YY}^{-1}$ for CCA,  $\Omegaz=\II$ for OPLS, and $\Omegaz=\II$ with  $\YY=\XX$ for PCA \cite{reinsel98,Sergio15}.


 

In order to extract uncorrelated features, the loss function \eqref{GOPLS_cost} is formally minimized subject to $\UU^\top\CC_{\XX\XX}\UU=\II$. However, it is proved in \cite{Sergio15} that the same solution is obtained constraining the minimization to $\WW^\top\Omegaz \WW=\II$. For the case in which $R(\UU)$ can be derived, it is possible to obtain a closed-form solution for $\UU$ as a function of $\WW$. Introducing this solution back into \eqref{GOPLS_cost}, the problem can be rewritten in terms of $\WW$ only. For instance, when $R\left(\UU\right)=||\UU||^2$ the solution for $\UU$ can be found by taking derivatives of \eqref{GOPLS_cost} with respect to $\UU$. After setting the result equal to zero, we obtain
\begin{equation}
\label{eq:U_solution}
\UU =\left(\CC_{\XX\XX}+\gamma \II\right)^{-1}\CC_{\XX\YY}\Omegaz \WW = \widetilde{\CC}_{\XX\XX}^{-1}\CC_{\XX\YY}\Omegaz \WW,
\end{equation}
where $\widetilde{\CC}_{\XX\XX}=\CC_{\XX\XX}+\gamma \II$. Now, replacing \eqref{eq:U_solution} into \eqref{GOPLS_cost}, the loss function can be written as a function $\WW$ only,
$${\cal L}(\WW)=\Trc\{\Omegaz \CC_{\YY\YY}\} - \Trc\{\WW^\top \Omegaz \CC_{\XX\YY}^\top\widetilde{\CC}_{\XX\XX}^{-1}\CC_{\XX\YY}\Omegaz \WW\}.$$
Minimizing this functional with respect to $\WW$, subject to $\WW^\top\Omegaz \WW=\II$, the solution is given in terms of the following generalized eigenvalue problem,
$$\Omegaz\CC_{\XX\YY}^\top \widetilde{\CC}_{\XX\XX}^{-1} \CC_{\XX\YY}\Omegaz \WW = \Omegaz \WW\Lambdaz,$$
which can be rewritten as a standard eigenvalue problem:
\begin{equation}
\label{eq:V}
\Omegaz^\frac{1}{2}\CC_{\XX\YY}^\top\widetilde{\CC}_{\XX\XX}^{-1}\CC_{\XX\YY}\Omegaz ^\frac{1}{2}\VV = \VV\Lambdaz,
\end{equation}
where we have defined $\WW=\Omegaz ^{-\frac{1}{2}}\VV$. Thus, $\UU$ can also be obtained as (see \eqref{eq:U_solution})
\begin{equation}
\label{eq:U}
\UU=\widetilde{\CC}_{\XX\XX}^{-1}\CC_{\XX\YY}\Omegaz^\frac{1}{2}\VV.
\end{equation}

Table \ref{Tab:summaryMVA} provides the above expression particularized for the CCA, OPLS and PCA methods. For each method, we show the corresponding eigenvalue problem that defines the solution for $\VV$, the associated $\WW$, and the solution for $\UU$ provided by \eqref{eq:U_solution}.
\begin{table}[h]
\caption{Summary of the most popular MVA methods: CCA, OPLS, and PCA.}
\label{Tab:summaryMVA}
\centering
\begin{tabular}{@{}llll@{}}
\toprule
& $\VV$ (eig. problem) & $\WW$ & $\UU$ \\
\midrule
CCA & $\CC_{\YY\YY}^{-\frac{1}{2}}\CC_{\XX\YY}^\top\widetilde{\CC}_{\XX\XX}^{-1}\CC_{\XX\YY}\CC_{\YY\YY}^{-\frac{1}{2}}\VV = \VV\Lambdaz$ & $\WW=\CC_{\YY\YY}^{\frac{1}{2}}\VV$ & $\widetilde{\CC}_{\XX\XX}^{-1}\CC_{\XX\YY}\CC_{\YY\YY}^{-\frac{1}{2}}\VV$ \\
OPLS & $\CC_{\XX\YY}^\top\widetilde{\CC}_{\XX\XX}^{-1}\CC_{\XX\YY}\VV = \VV\Lambdaz$ & $\WW=\VV$ & $\widetilde{\CC}_{\XX\XX}^{-1}\CC_{\XX\YY}\VV$\\
PCA & $\CC_{\XX\XX}^\top\widetilde{\CC}_{\XX\XX}^{-1}\CC_{\XX\XX}\VV = \VV\Lambdaz$ & $\WW=\VV=\UU$ & $\widetilde{\CC}_{\XX\XX}^{-1}\CC_{\XX\XX}\VV$ \\
\bottomrule
\end{tabular}
\end{table}




\subsection{Uncorrelation of the extracted features}
It is important to remark that, in the absence of regularization, the above approach still produces uncorrelated features, in spite of not enforcing it explicitly. To prove this, we set $\gamma = 0$ and multiply both sides of \eqref{eq:U} from the left by $\UU^\top\CC_{\XX\XX}$, arriving at:
\begin{equation}
\label{eq:UCxxU_UCxyW}
\UU^\top\CC_{\XX\XX}\UU=\UU^\top\CC_{\XX\YY}\Omegaz^\frac{1}{2}\VV.
\end{equation}
Next, substituting \eqref{eq:U} in \eqref{eq:V}, and premultiplying both sides by $\VV^\top$, we obtain
\begin{equation}
\label{eq:UCxyW_Lambda}
\UU^\top\CC_{\XX\YY}\Omegaz^\frac{1}{2}\VV=\Lambdaz.
\end{equation}
Therefore, by jointly considering \eqref{eq:UCxxU_UCxyW} and \eqref{eq:UCxyW_Lambda} we have
\begin{equation}
\label{eq:UCxyW_Lambda2}
\UU^\top\CC_{\XX\XX}\UU=\UU^\top\CC_{\XX\YY}\Omegaz^\frac{1}{2}\VV=\Lambdaz,
\end{equation}
which proves the uncorrelation of the extracted features, since $\Lambdaz$ is diagonal.

\section{Iterative solutions for regularized MVA methods}

In the case of non-derivable regularizations, the minimization of \eqref{GOPLS_cost} s.t. $\WW^\top\Omegaz \WW=\II$ has not a closed-form solution. This problem is found, for instance, when using LASSO regularization, or the very useful $\ell_{2,1}$-norm, that performs variable selection. In order to solve these regularized MVA methods, many authors have recurred in the literature to the following iterative coupled procedure:
\begin{enumerate}
\item Step-$\UU$. For fixed $\WW$ (satisfying $\WW^\top\Omegaz\WW=\II$), find the matrix $\UU$ that minimizes the following regularized least-squares problem,
\begin{equation}
\label{eq:reg_U}
\|\YY' - \UU^\top \XX\|_F^2 + \gamma R\left(\UU\right),
\end{equation}
where $\YY'=\WW^\top\Omegaz\YY$ is the transformed output data. Note that this step can take advantage of the great variety of existing efficient solutions for regularized least-squares problems \cite{Nie10,Grant08,Kim2008}.
\item Step-$\WW$. For fixed $\UU$, find the matrix $\WW$ that minimizes \eqref{GOPLS_cost} subject to $\WW^\top\Omegaz\WW = \II$ or, rewriting this step in terms of $\VV=\Omegaz ^{\frac{1}{2}}\WW$, solve $\VV$ by minimizing \begin{equation}\label{eq:procrustes}\|\bar\YY - \VV \XX'\|_F^2, \;\;\; \st \VV^\top\VV=\II,\end{equation}
where we have defined $\bar\YY=\Omegaz^{\frac{1}{2}}\YY$.
\end{enumerate}

%

Step-$\WW$ above is typically solved in the literature by using the orthogonal Procrustes approach. As we will see later, this solution neglects the uncorrelation among the extracted features and, despite of that, since it was initially proposed by \cite{Zou06} for the sparse PCA algorithm, it has been wrongly extended to supervised approaches such as sparse OPLS \cite{Gerven12}, group-lasso penalized OPLS (or SRRR) \cite{Chen12}, and $\ell_{2,1}$-regularized CCA (or L21SDA) \cite{Shi14}. Note that this Procrustes approach can still be considered mainstream, as it can be checked in the very recent works \cite{Lai16,Hu16}. An example of some other proposed Procrustes-based solutions can be found not only in theoretical proposals \cite{Qiao08,Qiao09,Dou10,Guo10,Han10,Liu14}, but also in real-world applications such as medical imaging \cite{Sjostrand06}, optical emission spectroscopy \cite{Ma08}, or decoding intracranial data \cite{Gerven12}.

Therefore, the main purpose of this paper is two-fold: (1) to alert the machine learning community about limitations of Procrustes when used as part of the above iterative method, as next section theoretically analyzes; (2) to propose an alternate method for the $\WW$-step that pursues feature uncorrelation, which is next presented (Section 3.2). 

\subsection{Generalized solution: $\WW$-step with Orthogonal Procrustes}

Problem \eqref{eq:procrustes} is known as Orthogonal Procrustes, whose optimal solution is given by $\VV_\text{P} = \QQ\PP^\top$, given the singular value decomposition $\CC_{\bar\YY\XX'}=\QQ\Sigmaz\PP^\top$ \cite{Schonemann1966}.

\subsection{Proposed solution: $\WW$-step as an eigenvalue problem}
\label{subsec:nuestro}

Using Lagrange multipliers, we reformulate \eqref{eq:procrustes} as the following maximization problem
$${\cal L}_\Xiz(\VV)=\Trc\{\UU^\top \CC_{\XX\YY}\Omegaz^{\frac{1}{2}} \VV\} - \Trc\{\left(\VV^\top\VV - \II\right)\Xiz\},$$
where $\Xiz$ is a matrix containing the Lagrange multipliers. Taking derivatives of ${\cal L}_\Xiz$ with respect to $\VV$, and setting this result to zero, we arrive at the following expression
\begin{equation}
\label{eq:UCxyW_Xi}
\UU^\top\CC_{\XX\YY}\Omegaz^\frac{1}{2}\VV=\Xiz.
\end{equation}

Now, since \eqref{eq:UCxyW_Lambda} needs to hold to guarantee uncorrelation of the extracted features, this implies that matrix $\Xiz$ should also be diagonal, which is not necessarily satisfied by the solution of \eqref{eq:procrustes}. In other words, when using the iterative procedure described above, it is not sufficient to impose $\VV^\top \VV = \II$ during the $\WW$-step, but we need to additionally impose \eqref{eq:UCxyW_Lambda} to get uncorrelated features.

Assuming that $\Xiz$ is a diagonal matrix, we can now premultiply both terms of \eqref{eq:UCxyW_Xi} by their transposes. Multiplying further by $\VV$ from the left, and using the fact that $\VV^\top \VV = \II$, we arrive at the following eigenvalue problem that is the basis of our method:
\begin{equation}
\label{eq:reg_V}
\Omegaz^{\frac{1}{2}}\CC_{\XX\YY}^\top\UU\UU^\top \CC_{\XX\YY}\Omegaz^{\frac{1}{2}}\VV=\VV\Xiz^2 = \VV\Lambdaz,
\end{equation}

Table 2 includes a summary of the $\UU$- and $\WW$-steps for the particular cases of regularized CCA, OPLS and PCA. Remember that $\WW$ can be straightforwardly computed from $\VV$ using the relations indicated in the last column of Table \ref{Tab:summaryMVA}.
\begin{table}[!t]
\caption{Proposed solution for the two coupled steps of most popular regularized MVA methods.}
\label{Tab:summary_regMVA}
\centering
\resizebox{\textwidth}{!}{
\begin{footnotesize}
\begin{tabular}{@{}lll@{}}
\toprule
& $\UU$-step (reg. LS) & $\WW$-step (eigenvalue problem)\\
\midrule

reg. CCA & $\displaystyle\argmin_\UU \|\YY' - \UU^\top \XX\|_F^2 + \gamma R\left(\UU\right)$ & $\CC_{\YY\YY}^{-\frac{1}{2}}\CC_{\XX\YY}^\top\UU\UU^\top \CC_{\XX\YY}\CC_{\YY\YY}^{-\frac{1}{2}}\VV = \VV\Lambdaz$\\
reg. OPLS & $\displaystyle\argmin_\UU \|\YY' - \UU^\top \XX\|_F^2 + \gamma R\left(\UU\right)$ & $\CC_{\XX\YY}^\top\UU\UU^\top \CC_{\XX\YY}\VV = \VV\Lambdaz$\\
reg. PCA & $\displaystyle\argmin_\UU \|\XX' - \UU^\top \XX\|_F^2 + \gamma R\left(\UU\right)$ & $\CC_{\XX\XX}^\top\UU\UU^\top \CC_{\XX\XX}\VV = \VV\Lambdaz$\\

\bottomrule
\end{tabular}\end{footnotesize}}

\end{table}

\subsection{Relationship between both solutions}

In this section we demonstrate that, in the absence of regularization, the solution to the eigenvalue problem \eqref{eq:reg_V} is given by $\VV_\text{EIG} = \QQ$, where the columns of $\QQ$ are the left singular vectors of matrix $\CC_{\bar\YY\XX'} = \QQ \Sigmaz \PP^\top$. This implies that the solution of our method is just a rotation of the solution obtained with Procrustes, $\VV_\text{P} = \QQ \PP^\top$. This rotation plays a crucial role at uncorrelating the extracted features. Indeed, in the experiments section we will see that not only more uncorrelated features can be obtained, but also that the extracted features are more effective at minimizing the overall objective function \eqref{GOPLS_cost}. 

We start by rewriting the singular value decomposition of $\CC_{\bar\YY\XX'}$ as
\begin{equation}
\label{eq:SVD}
\Omegaz^{\frac{1}{2}}\CC_{\XX\YY}^\top\UU = \QQ\Sigmaz\PP^\top,
\end{equation}
now, multiplying both terms of \eqref{eq:SVD} by their transposes from the right, we have
\begin{equation}
\label{eq:EVD}
\Omegaz^{\frac{1}{2}}\CC_{\XX\YY}^\top\UU\UU^\top \CC_{\XX\YY}\Omegaz^{\frac{1}{2}}=\QQ\Sigmaz^2\QQ^\top.
\end{equation}
Further multiplying both terms by $\QQ$ from the right, and comparing the result with \eqref{eq:reg_V}, we can see that the solution to the eigenvalue problem \eqref{eq:reg_V} is precisely $\VV_\text{EIG} = \QQ$ and $\Lambdaz = \Sigmaz^2$.

%

\section{Undesired properties of orthogonal Procrustes in regularized MVA}

In this section, we provide theoretical arguments about the unsuitability of using orthogonal Procrustes as the solution to the $\WW$-step, showing that the obtained solution lacks some desired  properties of MVA methods. In order to do so, we work on a generalization of the property declared in \cite{Zou06}, which states that a good regularized MVA method should reduce to the original (unregularized) MVA solution when the regularization term is suppressed. We will show that this is not the case when using the solution based on Procrustes. In particular, we study the two following issues that occur when setting ($\gamma=0$):
\begin{itemize}
\item The extracted features are not uncorrelated in general. This issue itself dismantles the correctness of all MVA methods based on the Procrustes solution.
\item Initialization of the iterative process becomes critical, and in some cases the algorithm may not progress at all (for $\gamma = 0$).
\end{itemize}

We demonstrate next the above statements, and discuss further on their implications.

\subsection{Uncorrelation of the input features using Procrustes}

Denoting the solution of the $\WW$-step as $\VV_\text{P}$ and the solution after the next $\UU$-step as $\UU_\text{P}$, we can use \eqref{eq:U_solution} to write
\begin{equation}
\UU_\text{P} = \CC_{\XX\XX}^{-1} \CC_{\XX\YY} \Omegaz^{\frac{1}{2}} \VV_\text{P},
\end{equation}
since this is the closed-form optimal solution of the $\UU$-step when regularization is removed. Now, it is easy to see that the autocorrelation matrix of the extracted features can be rewritten as
\begin{equation}
\label{ec:corr_procrustes}
\CC_{\XX'\XX'} = \UU_\text{P}^\top\CC_{\XX\XX}\UU_\text{P} = \VV_\text{P}^\top \Omegaz^{\frac{1}{2}}\CC_{\XX\YY}^\top \UU_\text{P}.
\end{equation}

Recalling that $\VV_\text{P} = \QQ \PP^\top$, and that $\CC_{\bar\YY\XX'} = \QQ \Sigmaz \PP^\top$, \eqref{ec:corr_procrustes} can be finally expressed as:
\begin{equation}
\CC_{\XX'\XX'} = \PP \QQ^\top \QQ \Sigmaz \PP^\top = \PP \Sigmaz \PP^\top,
\end{equation}
%
which is not diagonal in a general case, and, thus, there is no guarantee that the extracted features are uncorrelated. In fact, since $\Sigmaz$ is a diagonal matrix and $\PP$ is an orthogonal matrix ($\PP^\top=\PP^{-1}$), only permutations matrices P will result in uncorrelated features (diagonal $\CC_{\XX'\XX'}$); in this case, solutions $\VV_\text{P}=\QQ\PP^\top$ and $\VV_\text{EIG}=\QQ$ extract the same features, but not necessarily in the same order.



Experimental section will demonstrate that methods based on Procrustes do not necessarily enjoy the desired uncorrelation property and, even, when the regularization is cancelled (for $\gamma=0$) the correlation among the features will imply that part of the variance of the original data described by one feature will also affect other features. Furthermore, experiments we will show that, since this method does not explicitly pursue such uncorrelation, the obtained solution for $\gamma > 0$ results in higher correlation among the features than that of the  proposed  method.

\subsection{Proof of initialization dependency by applying orthogonal Procrustes approach}

In the experiments section, we will illustrate that the solution achieved when using Procrustes shows a significant variance when the initialization conditions are changed, even when the regularization term is removed. In this subsection, we pay attention to a particular issue associated to the initialization of the algorithm. In particular, we show that when $\VV$ is initialized with an orthogonal matrix (which is a quite common case in the literature) the algorithm does not progress at all.

Let $\VV^{(0)}$ denote an orthogonal matrix which is used for the algorithm initialization. Subsequently, $\UU^{(1)}$ and $\VV^{(1)}$ will denote the solutions to the $\UU$- and $\WW$-step, that can be obtained from $\VV^{(0)}$ as
%
\begin{enumerate}
\item $\UU^{(1)}=\CC_{\XX\XX}^{-1}\CC_{\XX\YY}\Omegaz^{-\frac{1}{2}}\VV^{(0)}$
\vspace{-.1cm}
\item $\Omegaz^{\frac{1}{2}}\CC_{\XX\YY}^\top\UU^{(1)} = \QQ\Sigmaz\PP^\top$
\vspace{-.1cm}
\item $\VV^{(1)}=\QQ\PP^\top$
\end{enumerate}

In order to express $\VV^{(1)}$ in terms of $\VV^{(0)}$, we use expressions for steps 1 and 2 to arrive at
\begin{equation}
\label{eq:CV_SVD}
\MM\VV^{(0)} = \QQ\Sigmaz\PP^\top,
\end{equation}
where we have defined $\MM =\Omegaz^{\frac{1}{2}}\CC_{\XX\YY}^\top {\CC}_{\XX\XX}^{-1}\CC_{\XX\YY}\Omegaz^{\frac{1}{2}}$ for compactness reasons.
Now, multiplying both sides of \eqref{eq:CV_SVD} by theirs transposes from the right and from the left, we obtain the following expressions (note $\MM$ is symmetric)
\begin{align}
\QQ\Sigmaz^2\QQ^\top &= \MM\VV^{(0)}\VV^{^\top\hspace{-0.05cm}(0)}\MM \\
\PP\Sigmaz^2\PP^\top &= \VV^{^\top\hspace{-0.05cm}(0)}\MM\MM\VV^{(0)}.
\end{align} 
From these, the following equalities that will be helpful for this demonstration are obtained:
\begin{eqnarray}
\label{eq:Q}
\QQ&=\MM\VV^{(0)}\VV^{^\top\hspace{-0.05cm}(0)}\MM\QQ\Sigmaz^{-2},\\
\label{eq:P}
\PP&=\VV^{^\top\hspace{-0.05cm}(0)}\MM\MM\VV^{(0)}\PP\Sigmaz^{-2}.
\end{eqnarray}

Finally, multiplying \eqref{eq:Q} by the transpose of \eqref{eq:P}, we can express $\VV^{(1)}$ as a function of $\VV^{(0)}$, and simplify the resulting expression as follows:
\begin{eqnarray*}
\VV^{(1)}&=& \MM\VV^{(0)}\VV^{^\top\hspace{-0.05cm}(0)}\MM(\QQ\Sigmaz^{-4}\PP^\top)\VV^{^\top\hspace{-0.05cm}(0)}\MM\MM\VV^{(0)}\\
&=& \MM\VV^{(0)}\VV^{^\top\hspace{-0.05cm}(0)}\MM(\MM\VV^{(0)})^{-4}\VV^{^\top\hspace{-0.05cm}(0)}\MM\MM\VV^{(0)}\\
  &=& \MM\MM\MM^{-4}\MM\MM\VV^{(0)} =  \VV^{(0)},
\end{eqnarray*}
where we made use of \eqref{eq:CV_SVD} and the fact that $\VV^{(0)}$ is orthogonal (i.e., $\VV^{^\top\hspace{-0.05cm}(0)} = \VV^{^{-1}\hspace{-0.05cm}(0)}$).

Therefore, we have proved that Procrustes based MVA iterative process results in its paralysis when the regularization term is canceled and $\VV$ is initialized as an orthogonal matrix. This is the case of the method proposed in \cite{Gerven12}, where the algorithm is initialized with the eigenvectors of $\CC_{\YY\YY}$. Note also that, since $\VV^\top\VV=\II$ (or $\WW^\top\Omegaz\WW=\II$) is imposed, an orthogonal matrix is a reasonable choice for initialization, the identity matrix being a classic choice in these cases.

\section{Experiments}
The previous section theoretically demonstrated the problems of Procrustes based MVA methods, as well as the validity of our proposal. In this section, we show empirically the differences of both approaches over a real problem. To that end, we are going to compare three implementations: iterative MVA solutions using Procrustes approach (referred to as ``Procrustes'') and the proposed solution (denoted ``Proposal''); furthermore, whenever possible, the original algorithm implementations (``Original'') will be included in the comparison. For all implementations, we are going to consider well-known MVA methods derived from the generalized framework: PCA, CCA and OPLS.

For this study, problem \textit{segment}  \cite{Blake98} will be used along this section. This dataset consists of 18 input variables, 7 output dimensions and 2390 samples. To be able to analyze initialization dependencies of iterative approaches, all results have been averaged over 50 random initializations. 

\begin{figure}[t]
  \centering
  \begin{tabular}{ccc}
     \includegraphics[width=4cm]{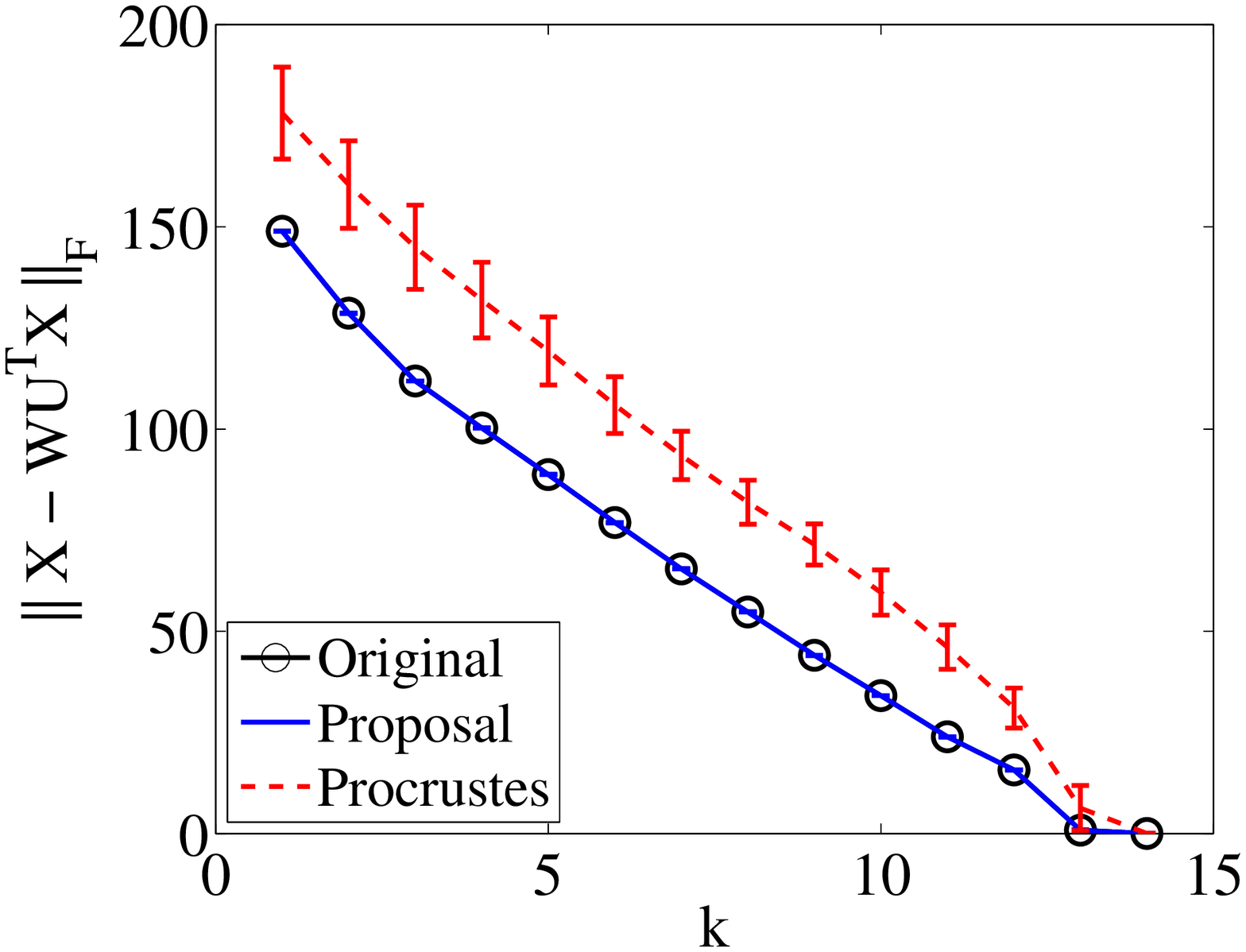}
     &
     \includegraphics[width=4cm]{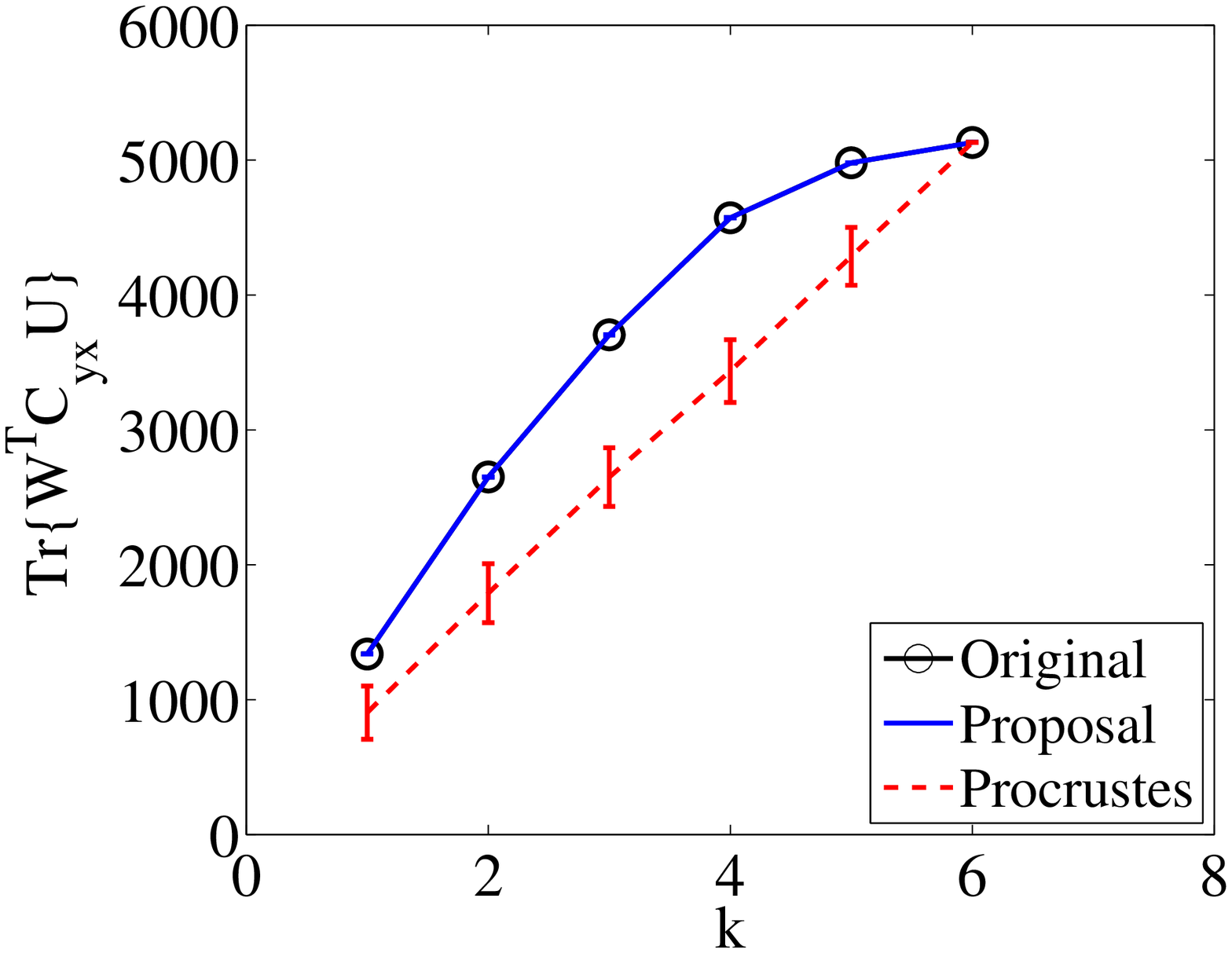}
     &
     \includegraphics[width=4cm]{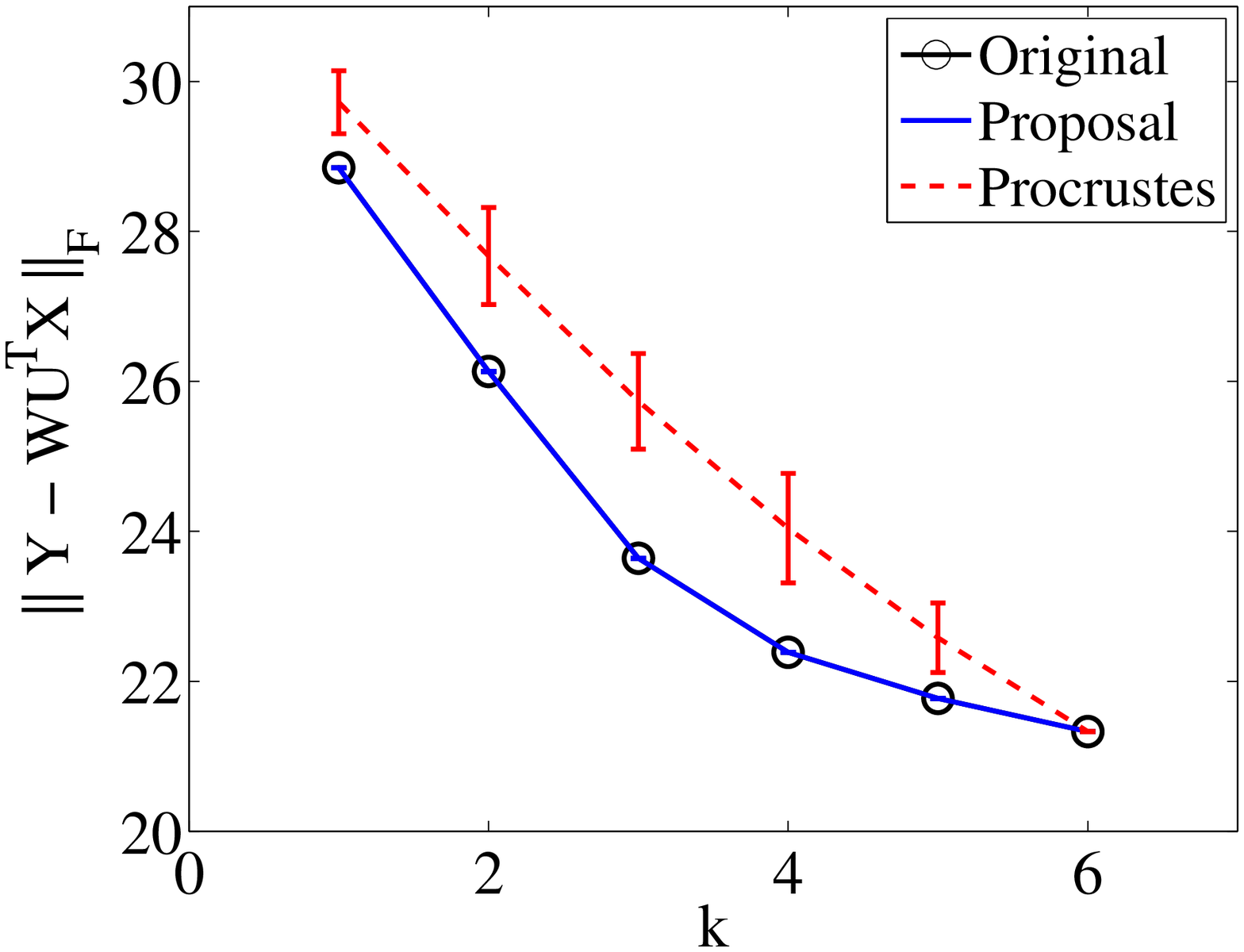}
     \\
     a) PCA & b) CCA & c) OPLS
 \end{tabular}
 \caption{Evolution of the global objective function with the number of extracted features ($k$) for the studied methods ($\gamma = 0$).}
 \label{fig:lossfunction}
\end{figure}

\begin{figure}[t]
  \centering
  \begin{tabular}{ccc}
     \includegraphics[width=4cm]{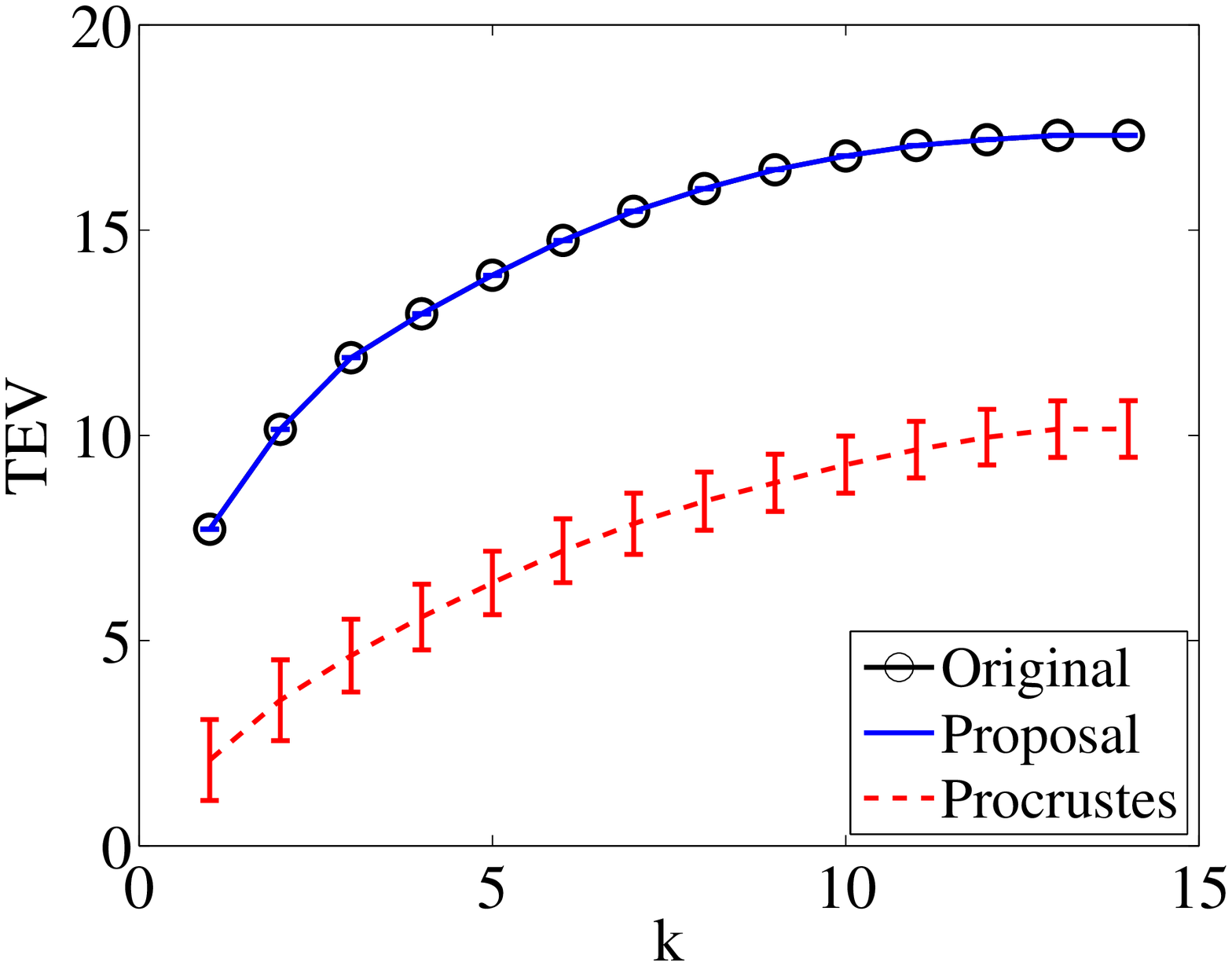}
     &
     \includegraphics[width=4cm]{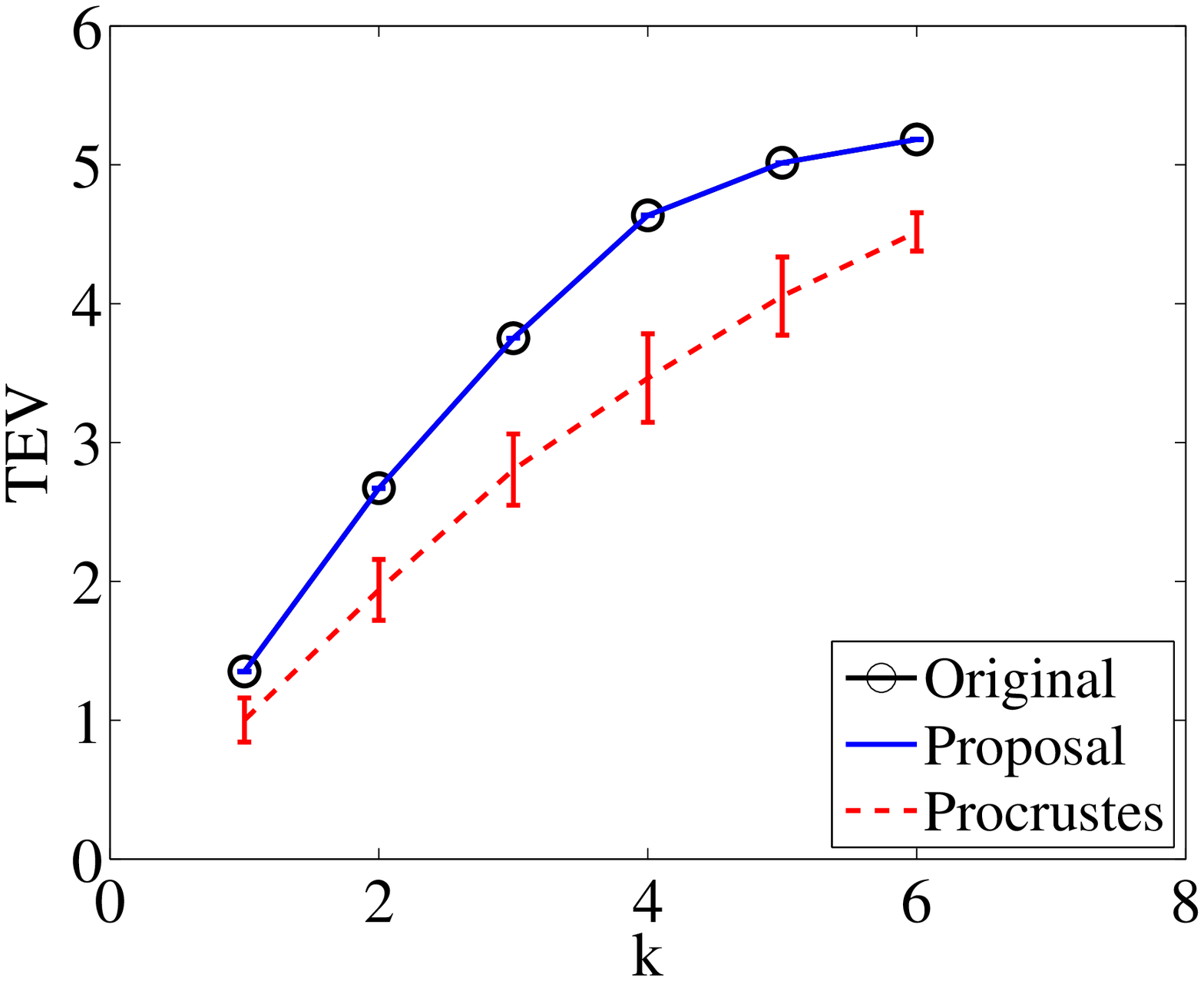}
     &
     \includegraphics[width=4cm]{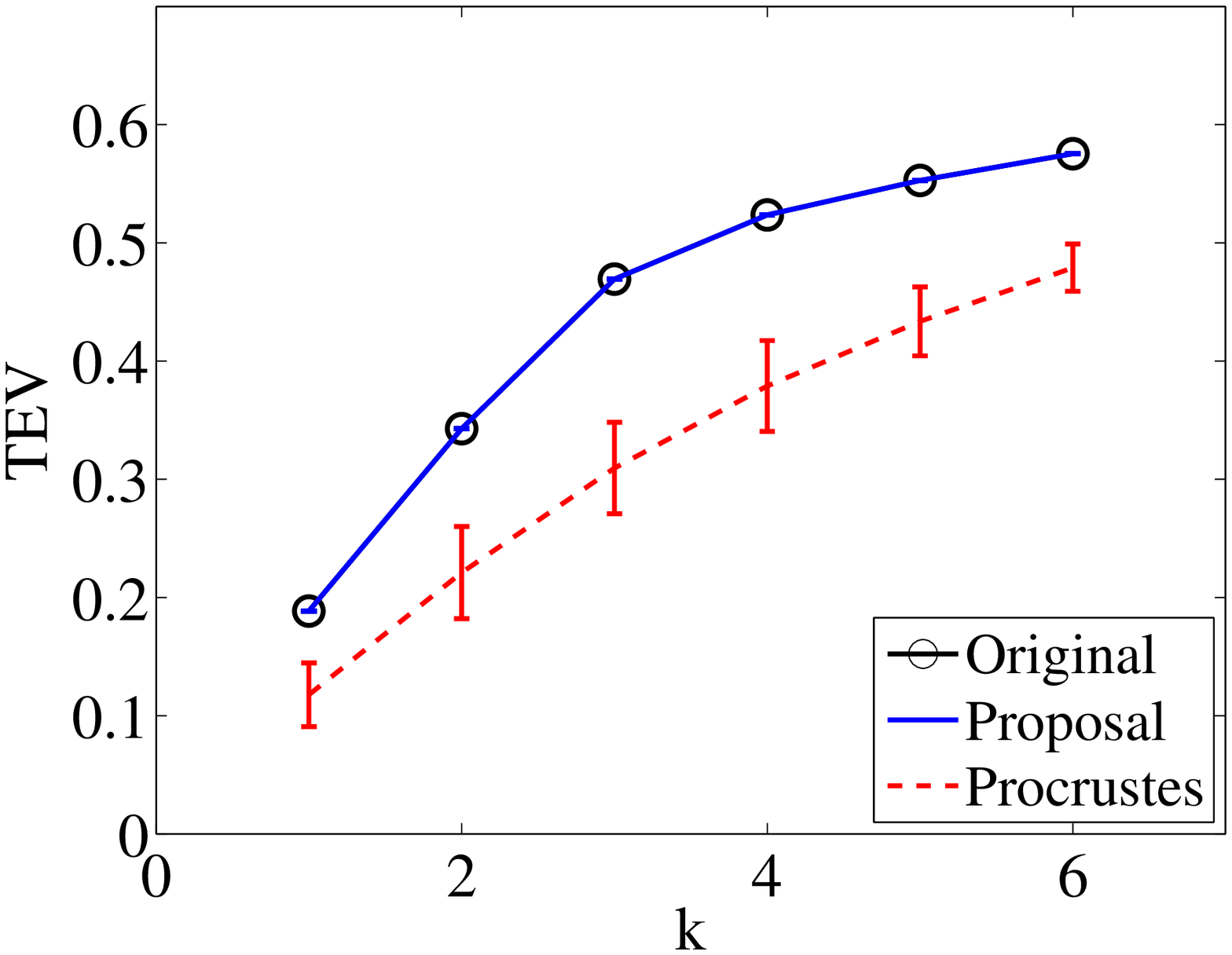}
     \\
     a) PCA & b) CCA & c) OPLS
 \end{tabular}
 \caption{TEV evolution with the number of features ($k$) for the methods under study ($\gamma = 0$).}
 \label{fig:variance}
\end{figure}

To start this evaluation, we are going to consider that no regularization is applied ($\gamma = 0$) and analyze the following algorithm behaviors when different number of extracted features are used:
\begin{enumerate}

\item {\bf Convergence to the minimum of the objective function}: evaluating the achieved value of the cost function \eqref{GOPLS_cost} \footnote{In CCA, we will consider its formulation as a maximization of a trace problem.}, we will be able to study whether the compared solutions are able to achieve the same performance as original MVA solutions (see Figure \ref{fig:lossfunction}). 
\item {\bf Information of the extracted variables}: we can measure when the extracted features are correlated or redundant by means of the {\it Total Explained Variance} (TEV) concept \cite{Zou06}, since its value decreases when there are relationships among features; thus, higher values of this parameter would indicate that the extracted features are more informative (see Figure \ref{fig:variance}). 
The  explained variance of a single variable would be given by computing the QR decomposition $\UU^\top\CC_{\XX\XX}\UU=\QQ\RR$, and taking the absolute value of the diagonal elements of R. Thus, \vspace{-.3cm}
$${\rm TEV}(k)= \sum_{j=1}^k{|\RR_{jj}|}.$$
\end{enumerate}

In the light of these results, we confirm those problems of the Procrusters based approaches, and check that the proposed MVA implementation can overcome them. In particular:
\begin{itemize}

\item From Figure \ref{fig:lossfunction}, we can conclude that, when $k<n_f$ extracted features are considered, Procrustes based approaches are not able to converge to the standard MVA solution. The proposed versions, however, converge to exactly the same solution as the original methods.

\item The Proposed MVA approach is able to extract more informative features. This is demonstrated by larger TEV values (for any value of $k$) than those achieved with the Procrustes method. This is a direct consequence of the uncorrelation among the extracted features.

\item Last but not least, standard deviation of all Procrustes based solutions reveal serious initialization dependency. Note that the proposed solutions, as well as the original MVA methods, converge to the same solution for all initializations.
\end{itemize}

\begin{figure}[t]
  \centering
  \begin{tabular}{ccc}
     \includegraphics[width=4cm]{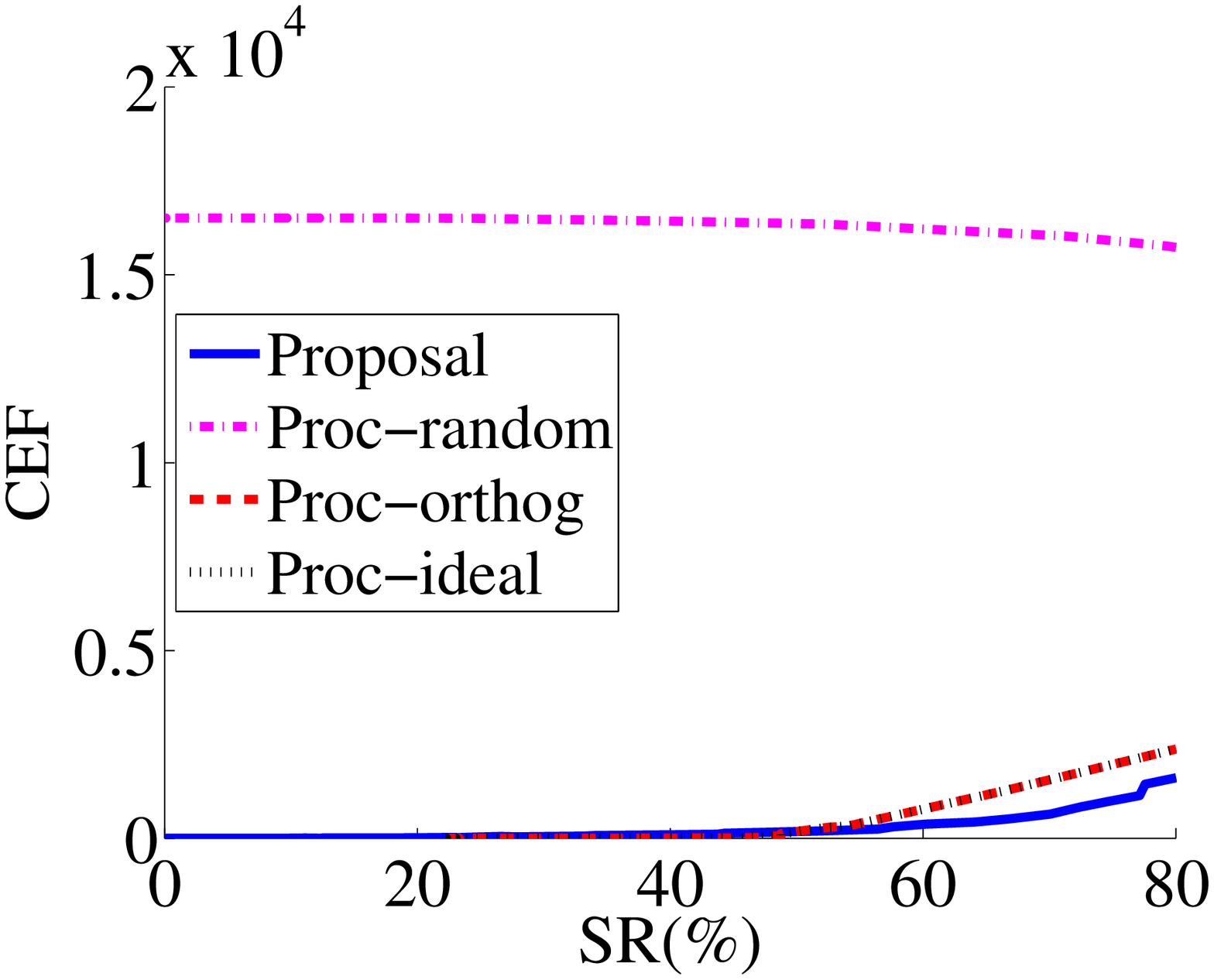}
     &
     \includegraphics[width=4cm]{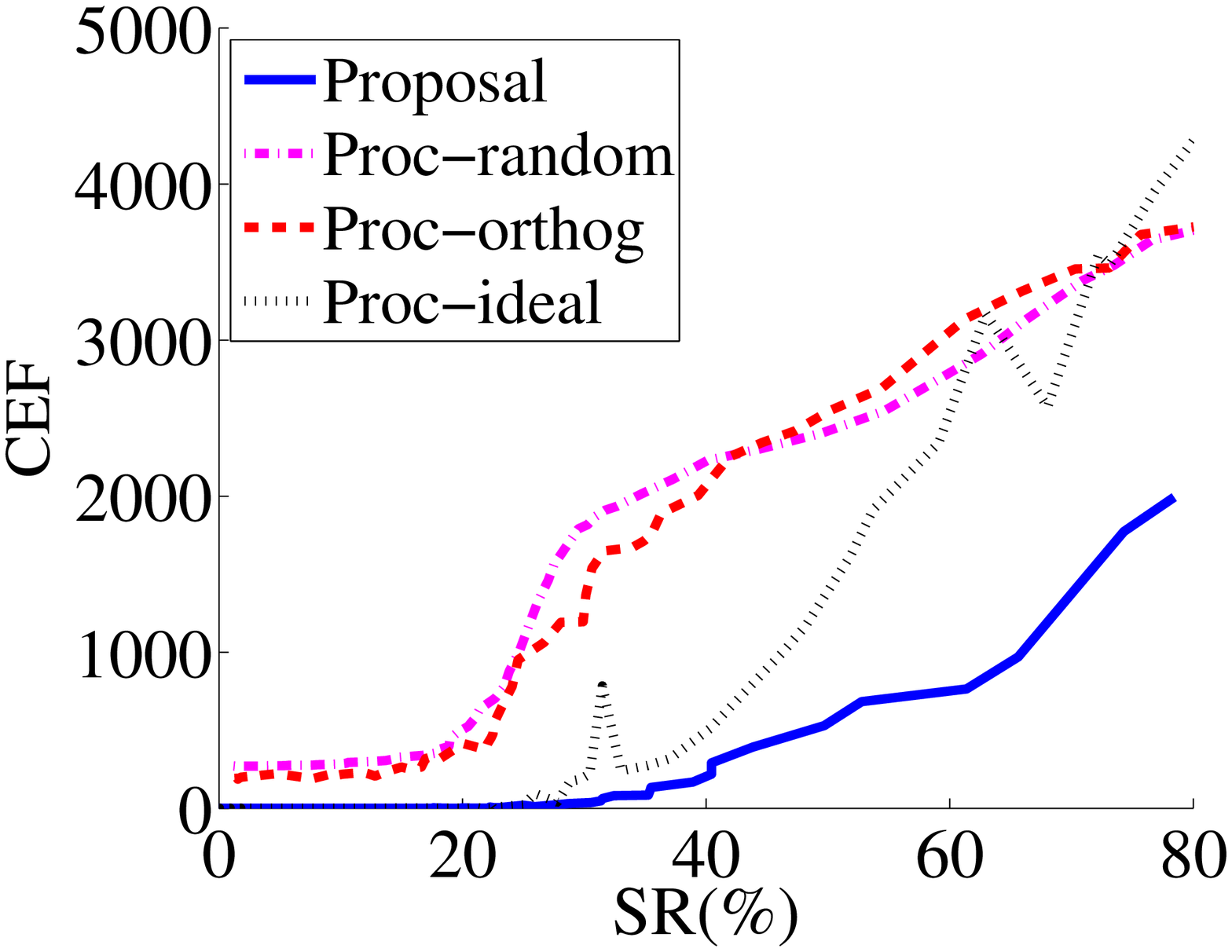}
     &
     \includegraphics[width=4cm]{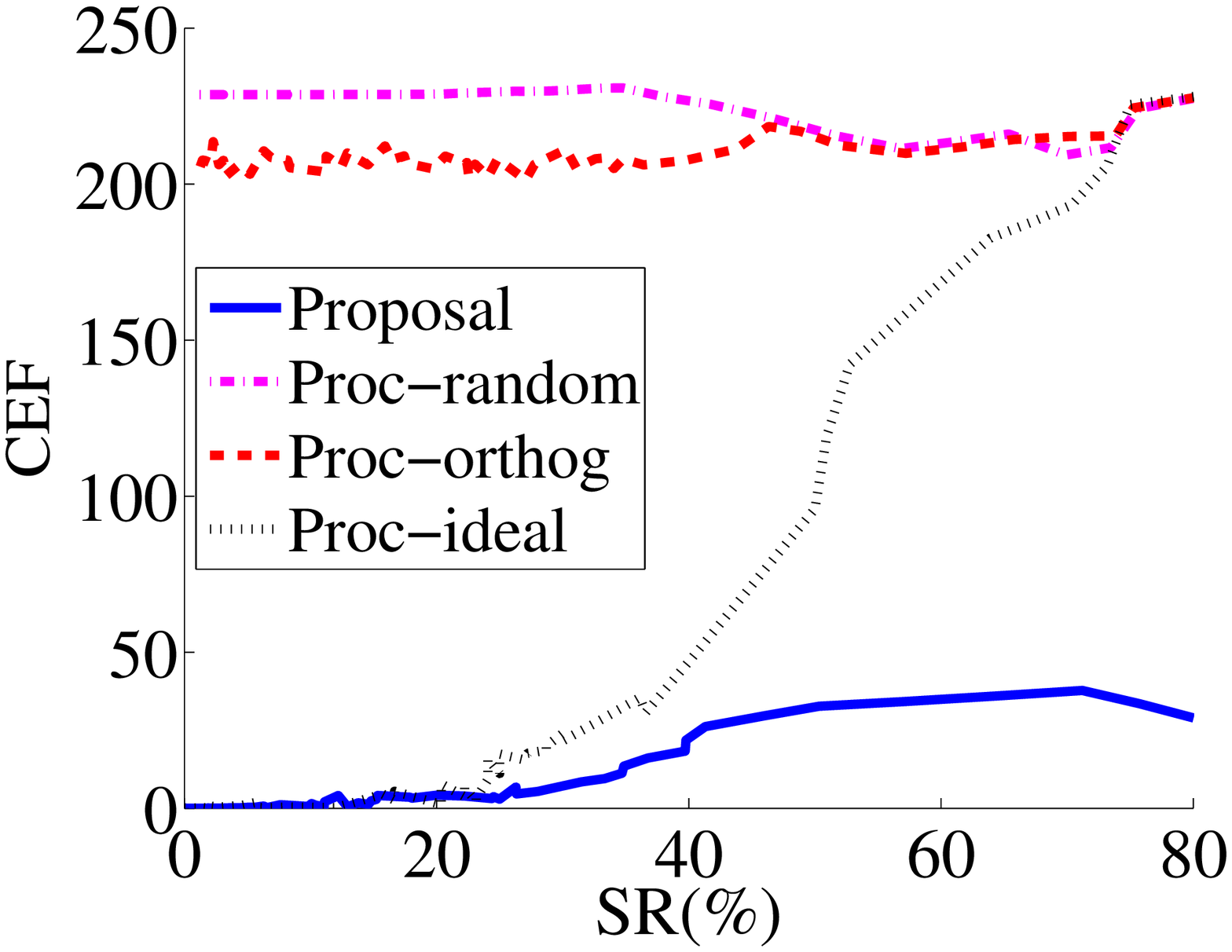}
     \\
     a) & b) & c)
 \end{tabular}
 \caption{CEF vs SR for the proposed MVA approaches and Procrustes based ones. Three different initializations are considered for all methods, as described in the text.}
 \label{fig:CEF}
\end{figure}

For the sake of completeness, we are also going to analyze uncorrelation among the extracted features when an $\ell_1$ penalty is used. For this purpose, we will directly measure the Correlation of the Extracted Features (CEF) by calculating the Frobenius norm 
$$ {\rm CEF}= {\Vert \UU^\top\CC_{\XX\XX}\UU - {\rm diag} \left( \UU^\top\CC_{\XX\XX}\UU \right)  \Vert }_F; $$
thus, CEF values different from zero would reveal correlations among the extracted components. In particular, Figure \ref{fig:CEF} includes the values of CEF parameter for different Sparsity Rates (SR) (from zero to 80\%\footnote{Take into account that exploring higher sparsity rates would lead all approaches to set $\UU = {\bf 0}$ and the CEF parameter would no longer make sense.}). For this study, three different initializations for Procrustes based methods are considered: (1) `Proc-random' that, as our proposal implementations, uses uniformly random values in the range from 0 to 1; (2) `Proc-orthog' which uses an orthogonal matrix given by the eigenvectors of $\CC_{\YY\YY}$ (or $\CC_{\XX\XX}$ in SPCA\footnote{Note that in this case SPCA initialization is equal to the standard PCA solution.}), as it is proposed \cite{Zou06} and \cite{Gerven12}; and, (3) `Proc-ideal' which directly starts the iterative process with the ideal solution when no sparsity regularization is used.

When the regularization parameter is included, all approaches (included the proposed ones) are not able to obtain an absolute uncorrelation among the extracted features; even so, CEF values reveal that our proposal gets a higher uncorrelation among the features than Procrustes approaches (independently of its initialization). Besides, when the regularization parameter is close to zero (SR = 0\%), Procrustes versions are only able to obtain uncorrelated features if they are initialized with the ideal solution (Proc-ideal). 
 

\section{Conclusions}

Solutions for regularized MVA approaches are based in an iterative approach consisting of two coupled steps. Whereas the first step easen the inclusion of regularization terms, the second  results in a constrained minimization problem which is generally solved as a orthogonal Procrustes problem. Despite the generalized use of this scheme, it fails in obtaining a new subspace of uncorrelated features, this being a desired property of MVA solutions.

In this paper we have analyzed the drawbacks of these schemes, proposing an alternative algorithm to force the uncorrelation property. The advantages of the proposed technique over the methods based on Procrustes have been discussed theoretically, and further confirmed via simulations.


\subsubsection*{Acknowledgments}
This work has been partly supported by MINECO project TEC2014-52289-R.

\bibliographystyle{IEEEtran}
\scriptsize
\bibliography{IEEEabrv,WhyHowAvoidProcrustes}

\end{document}